\documentclass[1p]{elsarticle}

\usepackage{adjustbox}
\usepackage{amssymb}
\usepackage{algorithm}
\usepackage[noend]{algorithmic}
\usepackage{float}
\usepackage{url}
\usepackage[numbers]{natbib}
\usepackage[figuresright]{rotating}

\usepackage{hyperref}

\newtheorem{definition}{Definition}
\newfloat{algorithm}{tbp}{loa}
\floatname{algorithm}{Algorithm}

\journal{Journal of Data \& Knowledge Engineering}

\begin{document}

\begin{frontmatter}
\title{Ultra-Fast Shapelets for Time Series Classification}

\author{Martin Wistuba\footnote{Corresponding author}}
\ead{wistuba@ismll.uni-hildesheim.de}
\author{Josif Grabocka}
\ead{josif@ismll.uni-hildesheim.de}
\author{Lars Schmidt-Thieme}
\ead{schmidt-thieme@ismll.uni-hildesheim.de}
\address{Information Systems \& Machine Learning Lab\\ University of Hildesheim\\ Marienburger Platz 22, 31141 Hildesheim, Germany}




\begin{abstract}
Time series shapelets are discriminative subsequences and their similarity to a time series can be used for time series classification.
Since the discovery of time series shapelets is costly in terms of
time, the applicability on long or multivariate time series is difficult.
In this work we propose Ultra-Fast Shapelets that uses a number of random shapelets. It is shown that Ultra-Fast Shapelets yield the same prediction quality as current state-of-the-art
shapelet-based time series classifiers that carefully select the shapelets by being by up to three orders of magnitudes.
Since this method allows a ultra-fast shapelet discovery, using shapelets for long multivariate time series classification becomes feasible.

A method for using shapelets for multivariate time series is proposed and Ultra-Fast Shapelets is proven to be successful in comparison to state-of-the-art multivariate time series classifiers on 15 multivariate time series datasets from various domains. Finally, time series derivatives that have proven to be useful for other time series classifiers are investigated for the shapelet-based classifiers. It is shown that they have a positive impact and that they are easy to integrate with a simple preprocessing step, without the need of adapting the shapelet discovery algorithm.
\end{abstract}


\end{frontmatter}


\section{Introduction}

Time series classification is a field of significant interest for
researchers because time series occur in various domains such as finance,
multimedia, medicine and more. A time series is a sequence of data
points that have a temporal relation between each other. For time
series classification it is common to identify motifs or local patterns
that have discrimination quality towards the target variable. One popular
method is to identify shapelets \cite{Ye2009}. Shapelets are discriminative
subsequences and have the property that the distance between a shapelet and its
best matching subsequence of a time series is a good predictor
for time series classification. Many methods try to find shapelets
and apply Shapelet Transformation \cite{Hills2014}. Shapelet Transformation
is the data transformation method that is used to convert the raw
time series data using the shapelets to a different data representation
that contains features that correspond to a specific shapelet and
its value is the minimal distance to the time series (see Section \ref{sub:Shapelet-Transformation-for-UTS}). The idea of
shapelets was mainly applied for univariate time series classification
but also for time series clustering \cite{Zakaria2012} and
early classification of multivariate time series \cite{Ghalwash2012}.

One of the biggest problems of shapelet discovery is that it is very
time-consuming. Hence, there are various methods of pruning the
candidate space \cite{Keogh2013}, improving the scoring function
that defines how good a shapelet is \cite{Mueen2011,Ye2009} or by
parallelization \cite{Chang2012}. Nevertheless, discovering shapelets
remains a slow procedure and is infeasible for large datasets.

\begin{figure}
\begin{centering}
\includegraphics[width=0.9\textwidth]{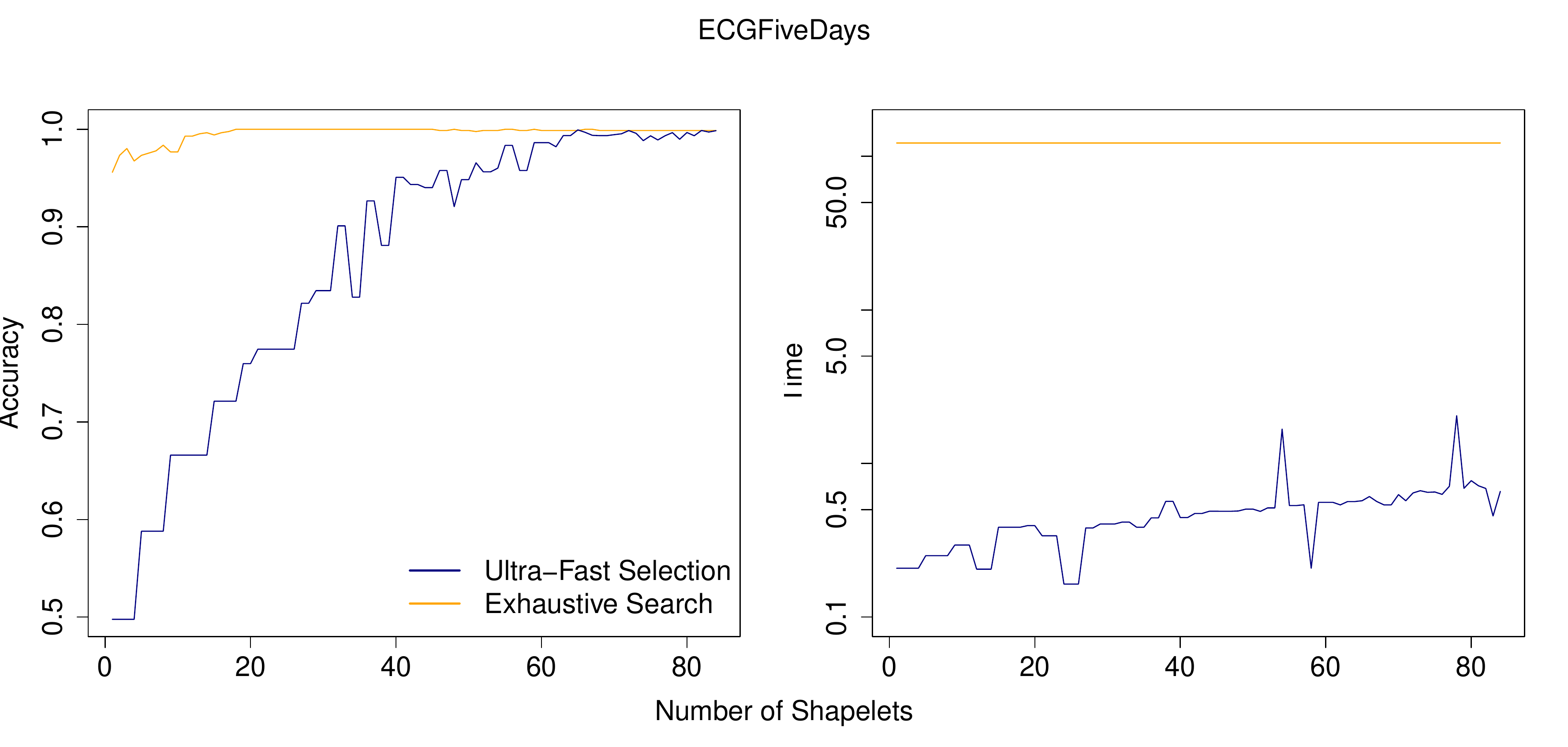}
\par\end{centering}
\caption{Comparing accuracy and runtime for exhaustively searching the best shapelets or selecting them at random.
The same accuracy can be achieved by randomly selecting shapelets but in a hundredth of the time.\label{fig:motivation}}
\end{figure}

We want to tackle the problem of slow shapelet discovery by introducing
an unsupervised method that does not need to compute the prediction accuracy of each single candidate.
Instead a pool of unsupervised, sampled shapelets is computed and a model is learned in the end.
The idea is to exploit the fact that discriminative motifs
are appearing frequently. This improves the shapelet discovery process and still a good accuracy can be achieved.
This effect is shown in Figure \ref{fig:motivation}.

The contributions are four-fold:
\begin{enumerate}
\item An ultra-fast way of extracting time series subsequences is proposed
that allows faster feature extraction than any method that tries
to identify discriminative subsequences in a supervised way but is
still comparable in terms of classification accuracy.
\item We propose a method which allows time series classification on multivariate
time series that concatenates features extracted on different streams.
This idea is evaluated on 15 datasets from various domains and compared
to state-of-the-art methods for multivariate time series classification.
\item Time series derivatives are considered as additional features for shapelet-based classifiers. Its
easy integration with a simple preprocessing step is demonstrated
as well as the positive impact on the classification accuracy for
shapelet-based classifiers.
\item A comparison between different shapelet-based methods using the original
authors' implementation is conducted. Additionally, the sensitivity for the hyperparameter that defines the number of shapelets
that are extracted is analyzed.
\end{enumerate}
The remainder of this paper is organized as follows. In the next section
related work is presented and delimited from our contributions. In
Section \ref{sec:Shapelet-Sampling-for-TSC}, the notation and problem
as well as a brief background description is provided. Also, our idea
of shapelet discovery is presented and motivated and finally extended
to multivariate time series classification. The section ends with
a discussion about time series derivatives. Afterwards, in Section
\ref{sec:Experiments}, it is presented that our idea of shapelet
discovery is faster than the state-of-the-art but nevertheless as
accurate for univariate time series. Finally, on 15 multivariate time
series datasets it is empirically proven that our method provides
better predictors for classification. The work is concluded in Section
\ref{sec:Conclusions}.

\section{Related Work\label{sec:Related-Work}}

The concept of shapelets, discriminative subsequences, was first introduced
by Ye et al. \cite{Ye2009}. The idea is to consider all subsequences
of the training data and assess them regarding a scoring function
to estimate how predictive they are with respect to the target. The
first proposed scoring function was information gain. Also other measures
like F-Stat, Kruskall-Wallis and Mood's median were considered \cite{Hills2014,Lines2012}.
It is possible to use the extracted subsequences to transform the
data and use an arbitrary classifier \cite{Hills2014}. Instead of
searching shapelets exhaustively, Grabocka et al. \cite{Grabocka2014}
try to learn optimal shapelets with respect to the target and report statistically
significant improvements in accuracy compared to other shapelet-based classifiers.

Shapelets have been used in many applications such as medicine \cite{Ghalwash2012},
gesture \cite{Hartmann2010} and gait recognition \cite{T.2012} and
even time series clustering \cite{Zakaria2012}.

Since a time series dataset usually contains many shapelet candidates,
a brute-force search is very time-consuming and hence many speed-up
techniques exist. On the one hand, there are smart implementations
using early abandon of distance computation and entropy pruning for
the information gain heuristic \cite{Ye2009}. On the other hand,
ideas to trade time for speed and reuse computations and to prune
the search space \cite{Mueen2011} as well as pruning candidates by
searching possibly interesting candidates on the SAX representation
\cite{Keogh2013} or using infrequent shapelets \cite{He2012} are
applied. Gordon et al. \cite{Gordon2012} learn a decision tree using random subsequences.
For each split they consider random subsequences and compute the information gain.
If for some time no better subsequence was chosen, the subsequence is used for the split.

In comparison to the related work, discriminative subsequences are not assessed
using a scoring function. Instead of that, subsequences are chosen at random
and subsequences that provide discriminative features are identified during
the learning process of a classifier. This leads to a faster feature extraction
process without much impact on the classification accuracy.
Our method is also not restricted to a specific classifier.

Shapelets are already considered for the multivariate time series
classification by Ghalwash et al. \cite{Ghalwash2012}. However, they
use multivariate instead of univariate shapelets and use them
for early classification. We follow a different
idea by using univariate shapelets that are specific for a stream
and by considering their interaction among each other using the classifier.
Hence, interaction between streams are learned and not assumed.

A very common method for multivariate time series classification
is to apply dimensionality reduction like singular value decomposition
on the data and then use any classifier on this data. This overcomes
the problem of time series with varying lengths \cite{Li2006,Weng2008}.
Other methods try to use similiarity-based methods that have
proven to be useful for univariate time series classification. For
example dynamic time warping was applied on multivariate time series
in the context of accelerometer-based gesture recognition \cite{Akl2010,Liu2009}.

Baydogan et al. \cite{Baydogan2014} use a symbolic representation
for multivariate time series. This is similar to SAX \cite{Lin2007}
but in contrast, the symbols are not fixed but learned in a supervised
way using random forests.

\section{Baselines\label{sub:Baselines}}

As it is later shown, Ultra-Fast Shapelets (UFS) is comparably accurate as any other
shapelet-based classifier but needs less time to extract the shapelets.
UFS is compared to three different methods to extract shapelets from
univariate data. Additionally, the reader will notice that the number
of hyperparameters needed for UFS is, in comparison, very low.

\subsection{Exhaustive Search (ES)}

The exhaustive search (ES) \cite{Hills2014,Mueen2011,Ye2009} considers
every subsequence in the training data and ranks it using a scoring
function $s$. As discussed in Section \ref{sub:Shapelet-Discovery-as-FSS},
this is equivalent to variable ranking. The scoring function $s$
is usually the information gain but also other quality measures like
Kruskal-Wallis, F-statistic and Mood's median were already considered.

As considering all candidates is infeasible for bigger datasets, the
candidates are reduced by considering only subsequences of specific
lenghts by choosing a minimum and a maximum length, sometimes a stepsize
greater than one. Obviously, as the length of the best subsequence
length is unknown, this are very sensitive hyperparameters. A further
hyperparamater is the number of shapelets that will be chosen. This
is also an important hyperparameter as setting it too low might lead
to not considering important features and setting it too high adds
to much noise for simple classifiers like Nearest Neighbor.

\subsection{Fast Shapelets (FS)}

Since exhaustively searching shapelets is very slow, the need for
a faster way for extracting shapelets exists. Hence, an approximative
method, so called Fast Shapelets (FS) \cite{Keogh2013}, was introduced.
The idea is reduce the dimension of the data by estimating
the SAX representation \cite{Lin2007} and searching on the reduced
space for features that are likely to be useful. Hence, mapping back
from the reduced space to the original space only few candidates are
left. The final features are estimated applying variable ranking.
Fast Shapelets is the fastest published method for shapelet
discovery we are aware of that yields comparable classification accuracy.

Like in the exhaustive search, there exist the hyperparameters to
prune the search space by subsequence length. However, it is also
possible to simply take all of them. Three new hyperparameters for
the dimensionality reduction are needed, i.e. window length, alphabet
size and word length, which might be less sensitive.

\subsection{Learning Shapelets (LS)}

Recently, a complete new idea was presented. Instead of restricting
the pool of possible candidates to those found in the training data
and simply searching them, Grabocka et al. \cite{Grabocka2014} propose
to consider the shapelets to be parameters that are optimized regarding
the loss as well. Hence, shapelets are not found using an approximative
measure for being useful for a model but directly optimized for it.
This means, it has an advantage because the method does not consider
a limited set of candidates but can choose arbitrary shapelets.
Disadvantages of this method are that its accuracy highly depends
on the initial shapelets and not every classification model can be
used. The number of hyperparameters is the same as the exhaustive
search but it will be shown that the running time can be
slower. In the following sections, this method is called Learning
Shapelets (LS).

\subsection{Symbolic Representation for Multivariate Timeseries}

Symbolic Representation for Multivariate Timeseries (SMTS) \cite{Baydogan2014}
is not based on shapelets but has its own way of generating features.
The idea is to represent time series as histograms of symbols where
a symbol represents a leaf of a decision tree. The raw data is transformed
such that for each time point $j$ of time series $T_{i}$ with label
$y_{i}$ the transformed data contains an instance $\left(j,t_{i,1,j},\ldots,t_{i,s,j},t_{i,2,j}-t_{i,1,j},\ldots,t_{i,m,j}-t_{i,m-1,j}\right)$
with label $y_{i}$. A random forest is trained on the transformed
data and its leafs are considered to be the symbols. The number of
occurences of a symbol in the raw data is counted and these symbol
histograms are used for the final classification step using random
forests.

\section{Ultra-Fast Shapelets for Univariate and Multivariate Time Series\label{sec:Shapelet-Sampling-for-TSC}}

Shapelets are often defined as discriminative subsequences and extracted
using a supervised quality measure. This process corresponds to a
feature subset selection \cite{Guyon2003} that is computational very
expensive as there usually exist many possible shapelet candidates.

In Section \ref{sub:Shapelet-Discovery-as-FSS} it will be shown that
most shapelet discovery methods are nothing else but feature subset
selection algorithms. An alternative way which is faster and is based
on feature sampling will be presented in Section \ref{sub:Shapelet-Discovery-as-FS}.
Subsequently, Section \ref{sub:Generalized-Shapelet-Sampling} introduces
a way of generating features from multivariate time series using subsequences.
Finally, Section \ref{sub:Derivatives-of-Time-Series} shows how to
integrate derivatives of time series to the proposed concept of Ultra-Fast Shapelets.

\subsection{Notation}

A univariate time series $T=\left(t_{1},\ldots,t_{m}\right)$ is a
sequence of $m$ data points, $t_{i}\in\mathbb{R}$, where $m$ is
called the length of the time series. For notational and illustratic
convenience, it is assumed that the length of each time series is
the same, although the presented methods can handle time series
of varying length.

A multivariate time series $T=\left(t_{1},\ldots,t_{m}\right)$ is
a sequence of $m$ vectors $t_{i}=\left(t_{i,1},\ldots,t_{i,s}\right)\in\mathbb{R}^{s}$
with $s$ streams where the time series $T_{j}=\left(t_{1,j},\ldots,t_{m,j}\right)$
is called a stream.

A dataset for univariate or multivariate time series classification
is a pair \\$\mathcal{T}=\left(\left(T_{1},\ldots,T_{n}\right)^{T},Y\right)$
where $Y=\left(y_{1},\ldots,y_{n}\right)\in\left\{ 1,\ldots,C\right\} ^{n}$
and the class of time series $T_{i}$ is $y_{i}$. The classification
task is now to use $\mathcal{T}$ to predict the correct classes for
further, unseen time series.

A shapelet for a dataset $\mathcal{T}$ is a time series of length
$l\leq m$ which is discriminative with respect to the target regarding
to a scoring function $s\ :\ span\left(\mathcal{T}\right)\times\mathbb{R}^{l}\rightarrow\mathbb{R}$,
where $s$ is usually the information gain.

\subsection{Shapelet Discovery as Feature Subset Selection\label{sub:Shapelet-Discovery-as-FSS}}

In this section the relationship between shapelet discovery and feature
subset selection is depicted. This relationship holds only for those
methods that choose subsequences as shapelets that occur in the
training data which is common in many methods \cite{Hills2014,Keogh2013,Lines2012,Mueen2011,Ye2009}.
For a better understanding, Shapelet Transformation is described in
Section \ref{sub:Shapelet-Transformation-for-UTS} and shapelet discovery
in Section \ref{sub:Shapelet-Discovery-is-Variable-Ranking}.

\subsubsection{Shapelet Transformation for Univariate Time Series\label{sub:Shapelet-Transformation-for-UTS}}

\begin{figure}
\begin{centering}
\includegraphics[width=\textwidth]{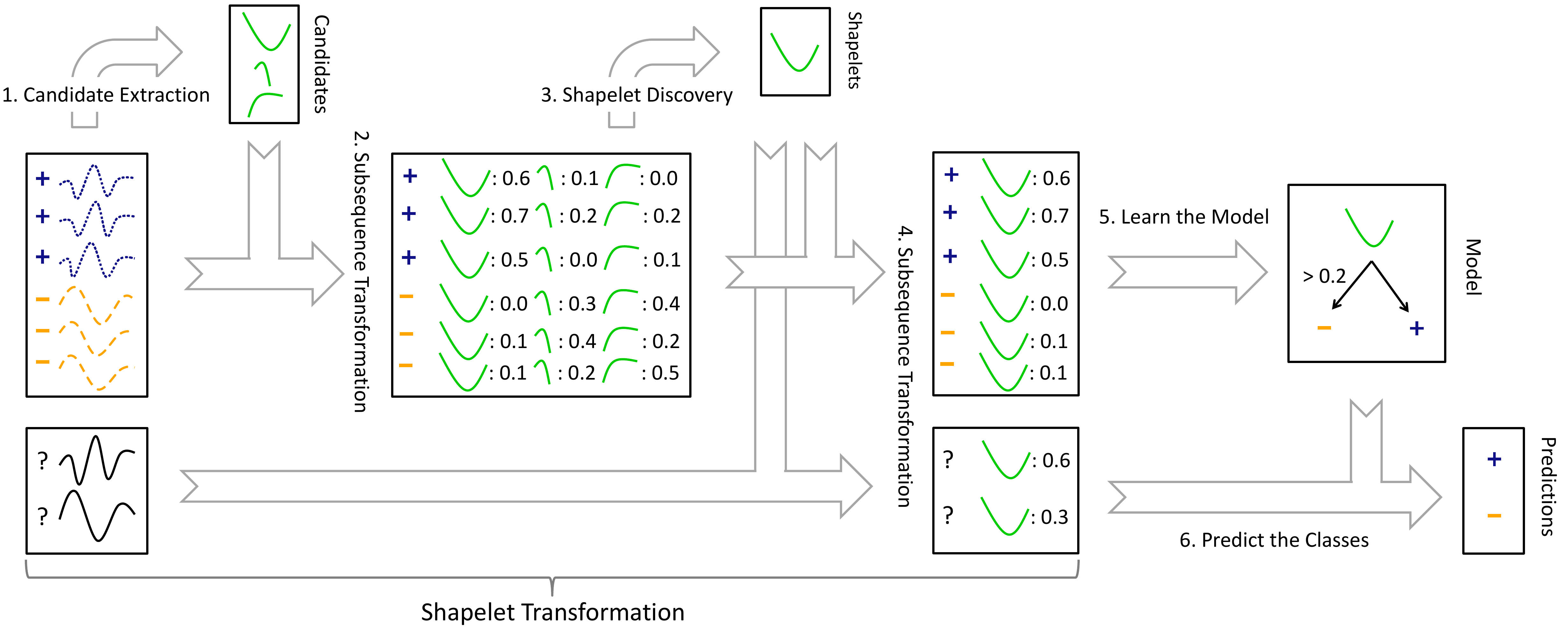}
\par\end{centering}

\caption{An illustration of the Shapelet Transformation process using a synthetically generated dataset.\label{fig:Shapelet-Transformation-for-UTS}}

\end{figure}

The term \emph{Shapelet Transformation} was introduced by Lines et
al. \cite{Lines2012}. Shapelet Transformation is a feature extraction
process which uses and extracts shapelets which are discriminative subsequences.
This idea is used for univariate time series classification latest
since Ye et al. \cite{Ye2009}. The extracted shapelets are used to
express the original time series dataset $\mathcal{T}=\left(\left(T_{1},\ldots,T_{n}\right)^{T},Y\right)$
as a shapelet transformed dataset $\mathcal{D}=\left(X,Y\right)$,
$X\in\mathbb{R}^{n\times p}$ which can be done in four steps (see
Figure \ref{fig:Shapelet-Transformation-for-UTS}):
\begin{enumerate}
\item \emph{Candidate Extraction:} From all time series $T_{1},\ldots,T_{n}$
subsequences $C_{1},\ldots,C_{q}$ are selected to be candidates.
There exist different methods to choose the candidates. Typically,
all subsequences of a specific length \cite{Hills2014,Mueen2011,Ye2009}
or that fulfill another informed criterion \cite{Keogh2013} are chosen.
\item \emph{Subsequence Transformation:} Using the $q$ candidates, the
raw data will be transformed. Each candidate $C=\left(c_{1},\ldots,c_{l}\right)$
is a predictor in the new representation and its value for a given
time series $T$ is the minimal Euclidean distance on the normalized
time series data i.e.
\begin{equation}
minNormDist\left(C,T\right)=\min_{i=1,\ldots,m-l+1}\left\{ \sqrt{\sum_{j=1}^{l}\left(c_{j}-\frac{t_{i+j}-\mu_{T}}{\sigma_{T}}\right)^{2}}\right\} \label{eq:minNormDist}
\end{equation}
where $\mu_{T}$ and $\sigma_{T}$ are the mean and
standard deviation of all data points $t_{i}$ of $T$, respectively.
Thus, the univariate time series dataset $\mathcal{T}$ can be transformed
into a dataset $\mathcal{D}'=\left(X',Y\right)$, $X'\in\mathbb{R}^{n\times q}$
with $x'_{i,j}=minNormDist\left(C_{j},T_{i}\right)$. Figure \ref{fig:minDist}
sketches the idea of the minimal distance. The minimal distance between
a subsequence and a time series is the Euclidean distance between
the subsequence and the best matching subsequence of the time series.
\begin{figure}
\begin{centering}
\includegraphics[height=4cm]{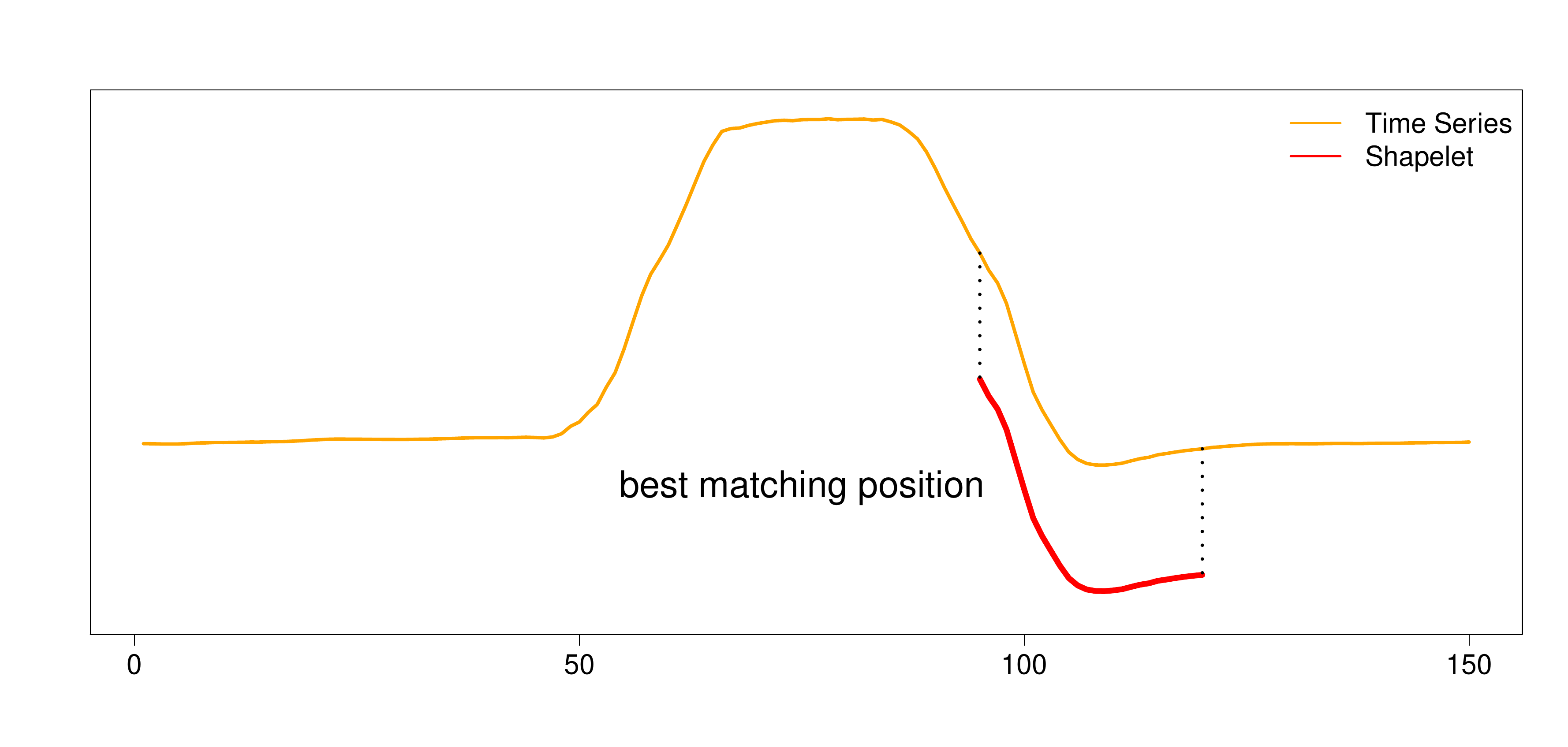}
\par\end{centering}

\caption{The minimum distance between a time series and a subsequence: Find
the best matching position and compute the Euclidean distance. The time series instance is taken from the GunPoint dataset.\label{fig:minDist}}

\end{figure}

\item \emph{Shapelet Discovery:} Using a scoring function $s$, all candidates
are ranked and the $p$ highest ranked subsequences $S_{1},\ldots,S_{p}$
are selected for further use. All the others are withdrawn.
\item \emph{Subsequence Transformation:} In the final step of the Shapelet
Transformation, the dataset $\mathcal{D}=\left(X,Y\right)$, $X\in\mathbb{R}^{n\times p}$
with $x_{i,j}=minNormDist\left(S_{j},T_{i}\right)$ is computed.
\end{enumerate}
After the completion of the Shapelet Transformation, there are no
restrictions whatsoever for a classifier.

\subsubsection{Shapelet Discovery is Variable Ranking\label{sub:Shapelet-Discovery-is-Variable-Ranking}}

The last section has shown how to apply Shapelet Transformation on
univariate time series. Now, the relation between shapelet discovery
and feature subset selection will be discussed. Therefor, the definition
that is used for a shapelet is recited here.
\begin{definition}
Given is a univariate time series dataset $\mathcal{T}$ and a scoring
function $s$ that ranks a subsequence of a time series $T$ according
to $\mathcal{T}$. Then, a shapelet $S$ is defined as a subsequence
of a time series $T\in\mathcal{T}$ which is among the $p$ highest
ranked subsequences regarding to the scoring function $s$.
\end{definition}
In the literature this kind of feature subset selection is called \emph{variable
ranking} and is a \emph{filter method} \cite{Guyon2003}. Filter
methods have the advantage of being computational fast and scalable
but have the disadvantages of choosing redundant features and ignore
dependencies among features and to the classifier. However, even though
variable ranking is one of the fastest feature selection methods,
it is still slow for shapelet discovery since the $\mathcal{O}\left(nm^{2}\right)$
features, where $n$ is the number of time series instances and $m$
is the length of a time series, are not given beforehand. First, the
minimal distance (Equation \ref{eq:minDist}) needs to be computed
before the quality can be assessed.

\subsection{Shapelet Discovery as Feature Sampling\label{sub:Shapelet-Discovery-as-FS}}

\begin{figure}
\begin{centering}
\includegraphics[width=\textwidth]{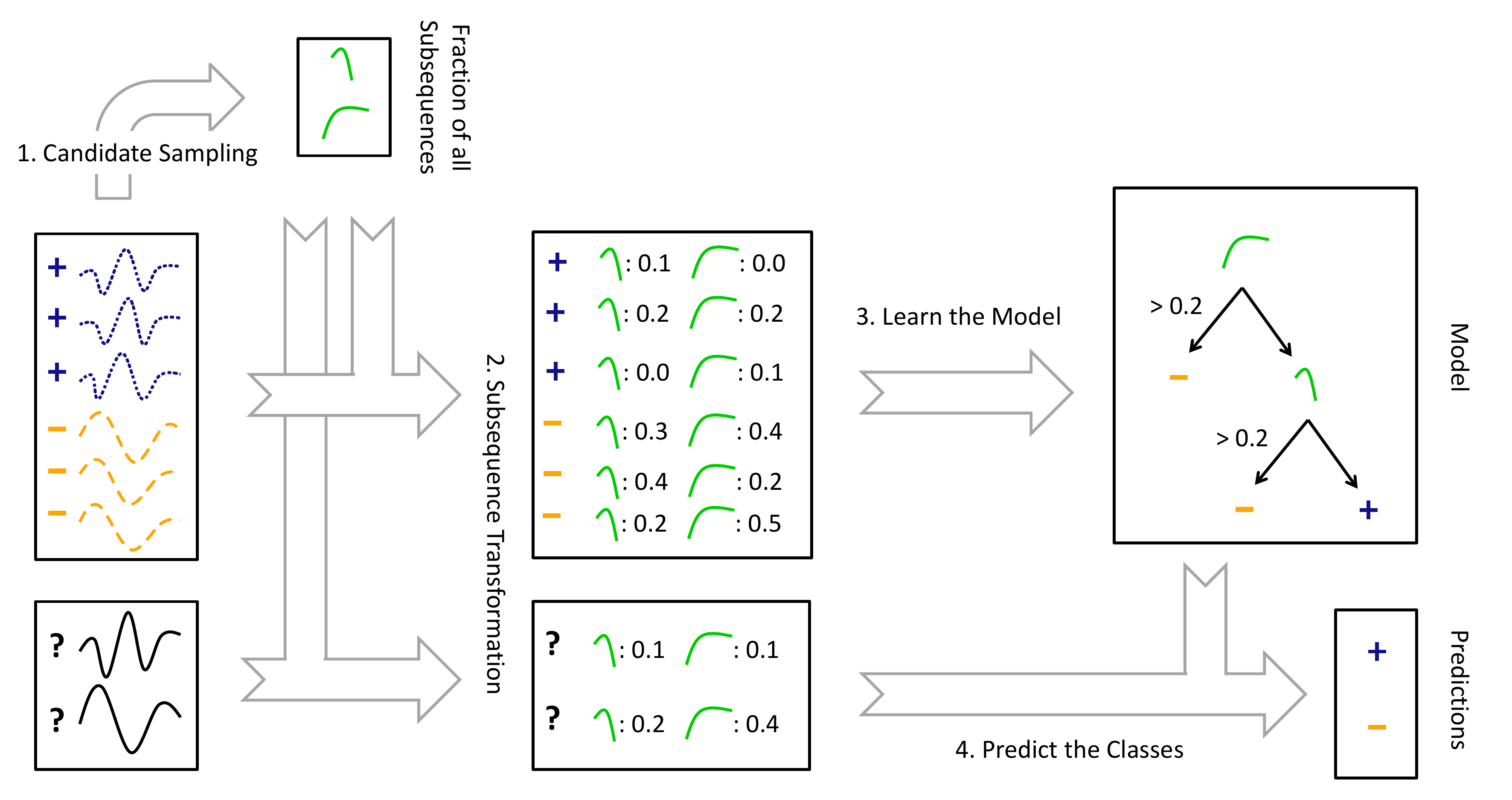}
\par\end{centering}

\caption{Ultra-Fast Shapelets for time series classification.\label{fig:Sampling-shapelets}}

\end{figure}

Here, a new way of shapelet discovery is proposed. Instead of considering
all candidates and applying subsequence transformation and variable
ranking (Shapelet Transformation, see Figure \ref{fig:Shapelet-Transformation-for-UTS}),
the predictors are chosen at random. This idea is sketched in Figure
\ref{fig:Sampling-shapelets}. Comparing Figures \ref{fig:Shapelet-Transformation-for-UTS}
and Figure \ref{fig:Sampling-shapelets}, one can see that the difference
is twofold. For the candidate extraction step a random subset of candidates
is chosen and steps 2 and 3 of the Shapelet Transformation, i.e. the
feature subset selection, are omitted. From now on, this way of selecting
subsequences will be called Ultra-Fast Shapelets.

\begin{figure}
\begin{centering}
\includegraphics[height=5cm]{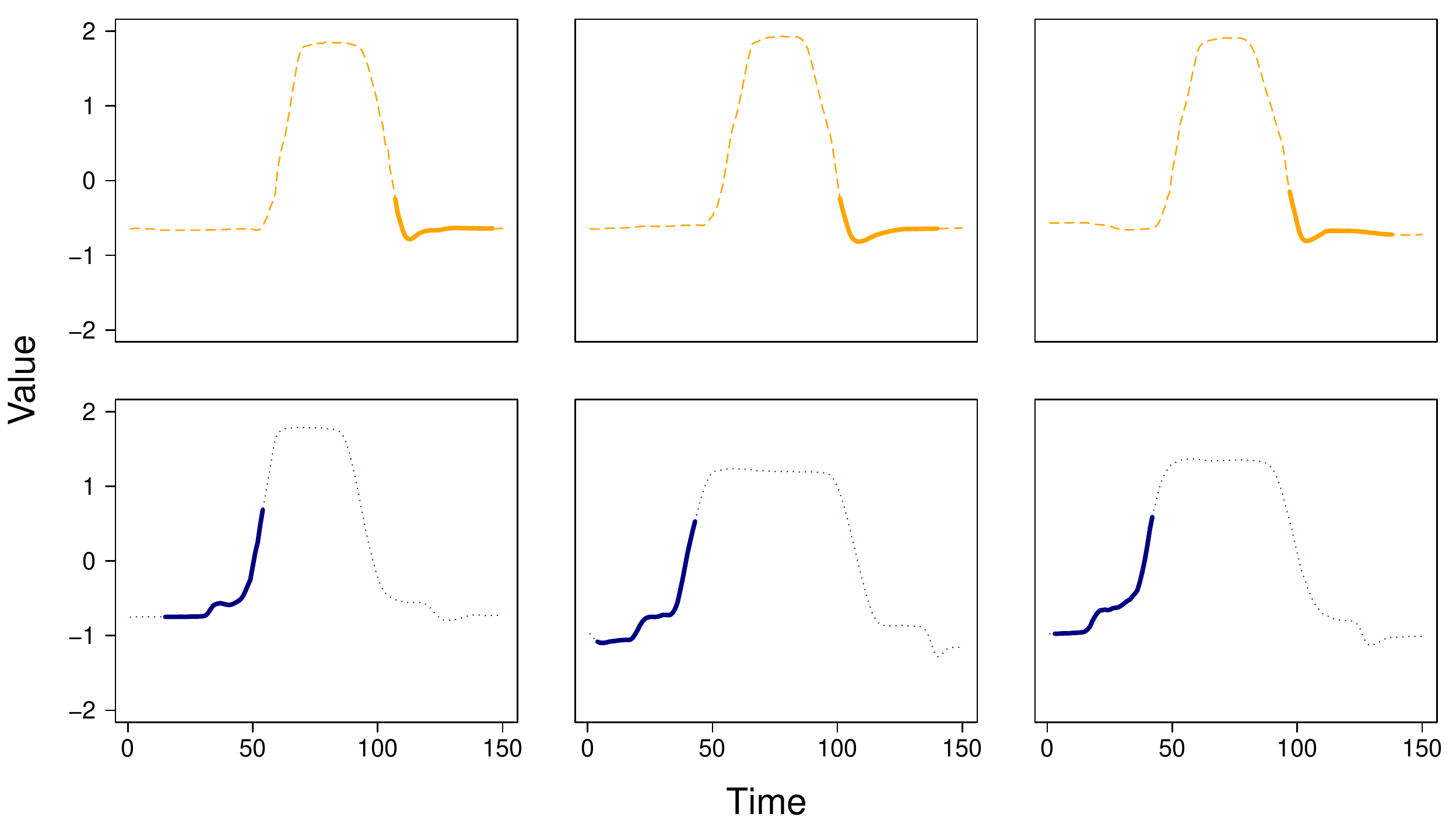}\hfill{}
\includegraphics[bb=50bp 0bp 354bp 550bp,clip,height=5cm]{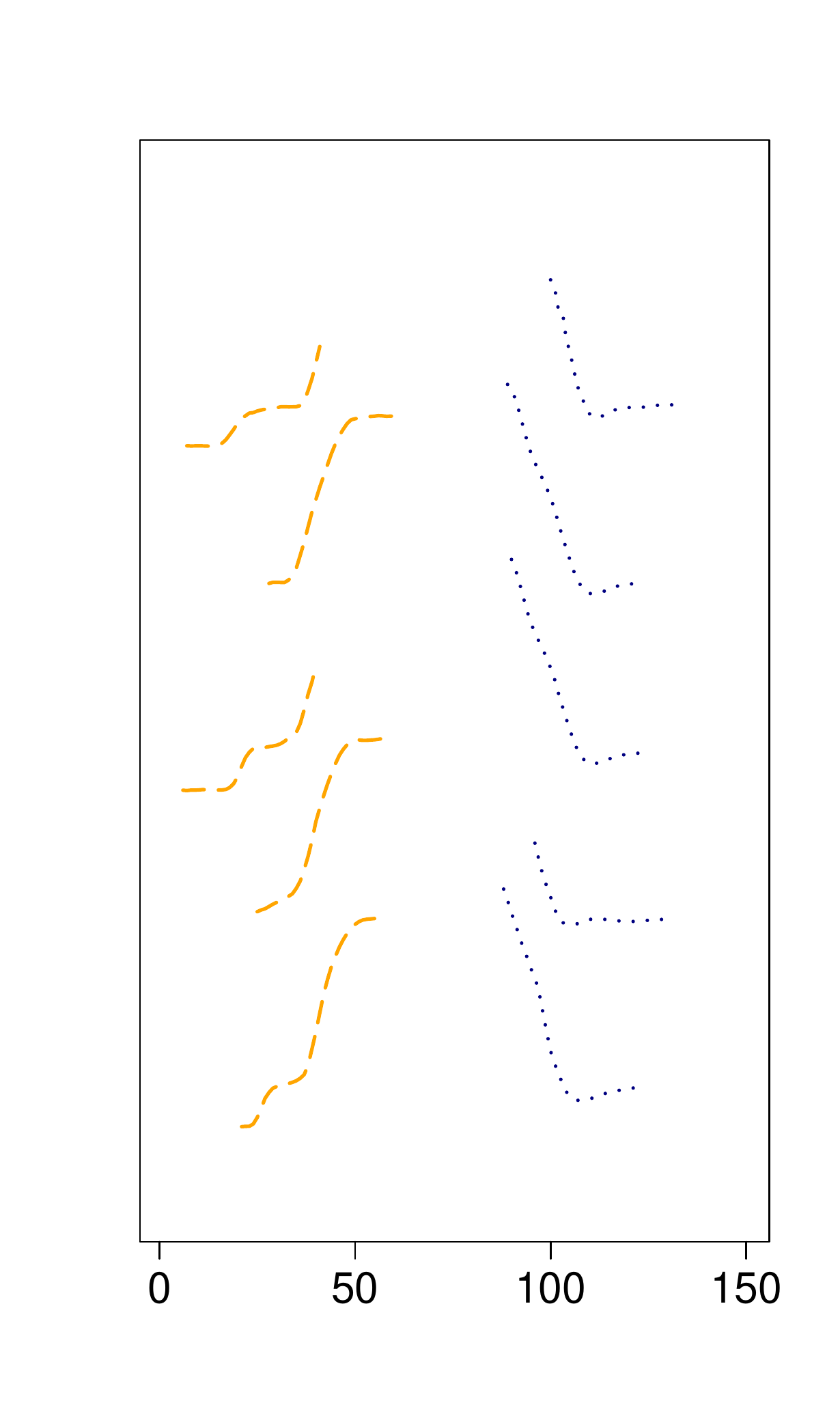}
\par\end{centering}

\caption{Left: Six random instances from the GunPoint dataset. The task is
to decide whether a person is drawing a gun or pointing at something.
The dashed, orange plots are showing examples for the pointing class,
the dotted, blue plots are showing examples for the gun drawing class.
Subsequences that are supposed to be discriminative for a class are
highlighted. Right: A selected subset of randomly chosen subsequences
of length 35 to 45. The dashed, orange subsequences are chosen from
the pointing class, starting before time point 40. The dotted, blue
subsequences are chosen from the drawing gun class, starting between
time points 80 and 100.\label{fig:Gun_Point}}

\end{figure}

The GunPoint dataset will be used to motivate the idea of Ultra-Fast Shapelets.
The GunPoint dataset is an activity recognition dataset
where the task is to distinguish whether a human being is lifting
his arm to point at something or whether he is drawing a weapon. Example
instances of this dataset are shown in Figure \ref{fig:Gun_Point}.
This figure also shows the shapelets found using the variable ranking
method which are considered to be useful for classification \cite{Hills2014}.
First, the question is whether similar good subsequences can be found
if randomly subsequences of arbitrary length are considered and second,
if useless subsequences are a problem. Lets stick to the GunPoint
dataset and assume that there are only the two clusters of shapelets
shown in Figure \ref{fig:Gun_Point}. Obviously, it is enough to find
just one representative for each shapelet cluster since they are redundant
and do not give further information. Since they are typical for the
data, it is assumed that one out of the two shapelets is found in
each time series. The GunPoint dataset contains 50 training instances
and hence approximately 50 shapelets. For this dataset $1\%$ of the
subsequences are chosen at random. Hence, the probability for finding
a representative for both shapelets clusters is quite unlikely. Nevertheless,
subsequences within the same interval as one of the shapelets which
are a bit longer or shorter are obviously also very good predictors.

In the example, all shapelets are of length 40 but one can assume
that subsequences of lengths between 35 and 45 in the same time interval
have similar predictive quality. Thus, there are not 50 candidates
but 1,230. If one agrees that this holds, the probability of selecting
a subsequence which is very close to one of the shapelets discovered
by variable ranking is very high. At this point, the reader may have
already noticed that there are further discriminative parts that might
have not been considered by the variable ranking. Since only the best
$p$ features are considered in the variable ranking method and
since there are many similar and hence equally high ranked features,
the effective number of features in the end can be very small. If
$p$ is not chosen high enough, discriminative parts like in the dashed,
orange plots around time 50 and in the dotted, blue plots around time
110 (Figure \ref{fig:Gun_Point}) are not considered.

\begin{figure}
\begin{centering}
\includegraphics[width=0.9\textwidth]{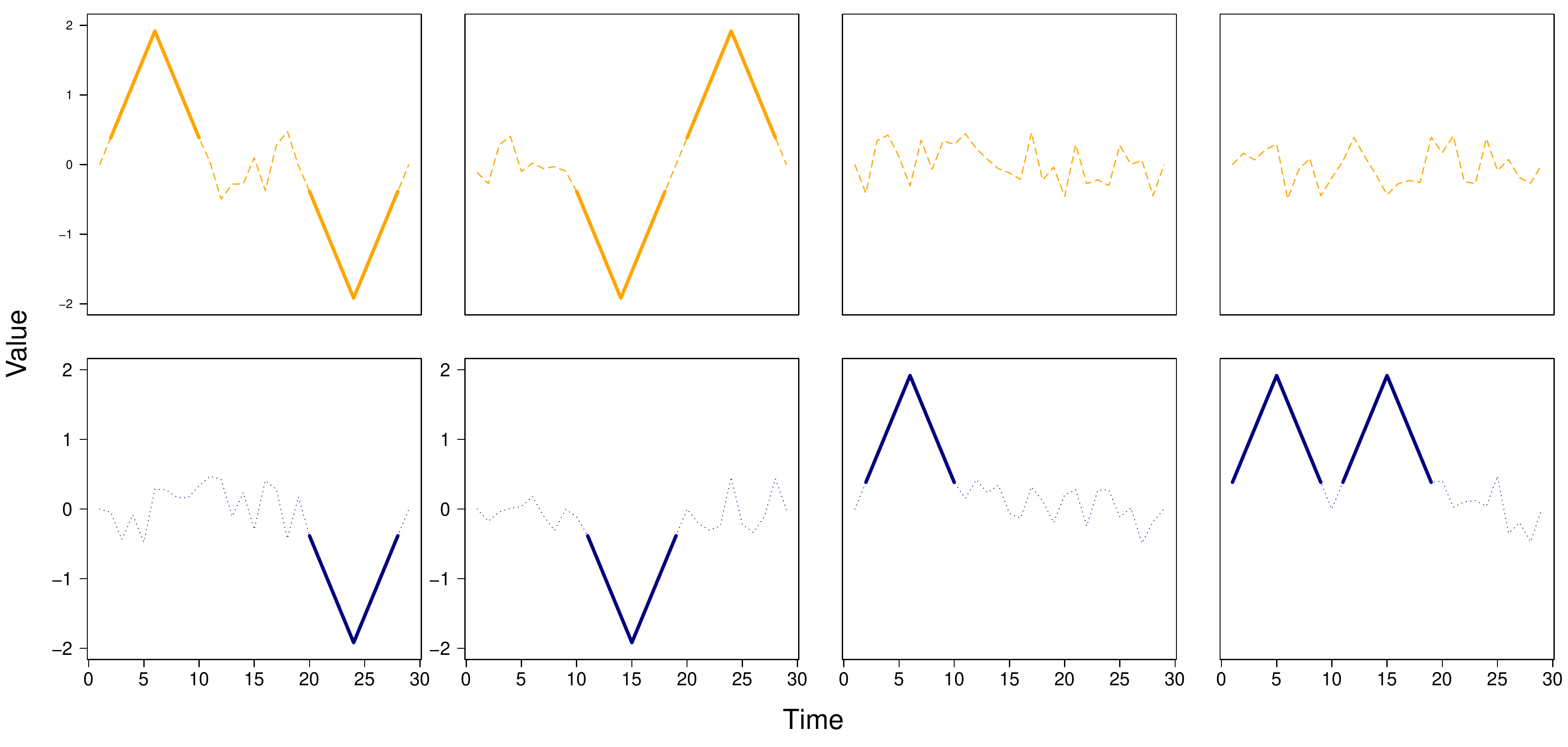}
\par\end{centering}

\caption{A synthetic dataset where the combination of two subsequences is needed
to predict the correct class. The important subsequences can be at
any point and arbitrary often.\label{fig:feature-interaction}}

\end{figure}

Finally, it is more likely that Ultra-Fast Shapelets finds interacting
features than variable ranking. Lets think about a dataset with two
classes (Figure \ref{fig:feature-interaction}). The one class has
either only noise or at least one $\vee$-shaped and one $\wedge$-shaped
subsequence which are interrupted by arbitrary long subsequences of
noise. The other class either has only subsequences of type $\vee$
divided by subsequences of noise and none of type $\wedge$ or vice
versa. If features are extracted by variable ranking, the subsequences
$\vee$ and $\wedge$ will have a bad score since the accuracy of
each of them alone is about $50\%$ which is as good as random. Therefor,
they will not be selected and no good classifier can be trained. On
the other hand, Ultra-Fast Shapelets will choose some of them as discussed
before and a perfect classifier can be trained. 

Concluding, time series contain many duplicates of subsequences where
some of them are useful and some are not. This motivates the idea
of randomly selecting subsequences. A well regularized classifier
can then be used to identify useful subsequences and their interactions.

\subsection{Generalized Ultra-Fast Shapelets\label{sub:Generalized-Shapelet-Sampling}}

This section describes how Ultra-Fast Shapelets can be generalized such
that the implementation can be used for univariate and multivariate
time series classification. The proposition is to transform multivariate
time series using shapelets in a similar way as described for the
univariate case before. In the first step, shapelets are chosen at
random from each of the $s$ streams such that sets of shapelets $\mathcal{S}_{1},\ldots\mathcal{S}_{s}$
are extracted. Ignoring temporal relations between the streams, the
raw data is transformed into a dataset $\mathcal{D}$ such that the
features are generated on each stream as in the univariate case and
then concatinated. Formally, using the sets $\mathcal{S}_{i}$ of
size $\frac{p}{s}$, the dataset $\mathcal{D}=\left(X,Y\right)$,
$X\in\mathbb{R}^{n\times p}$ is computed where $x_{i,j}=minDist\left(S_{j-\left(\left\lceil j\frac{s}{p}\right\rceil -1\right)\frac{p}{s}},T_{i,\left\lceil j\frac{s}{p}\right\rceil }\right)$
and $S_{j-\left(\left\lceil j\frac{s}{p}\right\rceil -1\right)\frac{p}{s}}\in\mathcal{S}_{\left\lceil j\frac{s}{p}\right\rceil }$.

Algorithm \ref{alg:Generalized-Shapelet-Sampling} explains this process
in detail. In line 2, the ratio $f$ between the number of features
$p$ and the number of all subsequences of arbitrary length is computed.
In lines 4-9, subsequences of different length are extracted uniformly
at random. After extracting them, the subsequence-transformed dataset
is estimated by computing the distance of a subsequence to each time
series in lines 12-16. For the distance function in line 16 we have chosen one of
the distance functions defined in Equations \ref{eq:minNormDist}
and \ref{eq:minDist}.

For the experiments on the multivariate datasets, a different distance
function between a shapelet and a time series is considered because it provides
better results for some datasets: the minimal
Euclidean distance between a shapelet and the unnormalized time series
\begin{equation}
minDist\left(S,T\right)=\min_{i=1,\ldots,m-l+1}\left\{ \sqrt{\sum_{j=1}^{l}\left(s_{j}-t_{i+j}\right)^{2}}\right\} .\label{eq:minDist}
\end{equation}

\begin{algorithm}
\caption{Generalized Ultra-Fast Shapelets}
\begin{algorithmic}[1]
\label{alg:Generalized-Shapelet-Sampling}
\REQUIRE{Dataset $\mathcal{T}=\left(\left(T_1,\ldots,T_n\right)^T,Y\right),\ T_{i}\in\mathbb{R}^{m\times s}$, number of features $p$, $c_1,\ldots,c_m$ where $c_i$ is the number of all subsequences of length $i$}
\ENSURE{Subsequence transformed dataset $\left(X,Y\right)$}
\STATE $\triangleright$ Choose $p$ subsequences at random
\STATE $f \leftarrow p\left(\sum_{l=3}^{m}c_{l}\right)^{-1}$
\STATE subsequences $\leftarrow \emptyset$
\FOR{$l=3$ to $m$}
  \FOR{$k=1$ to $s$}
    \FOR{$round\left(f\cdot c_{l}\right)$ times}
      \STATE $i\leftarrow\mathcal{U}\left(1,n\right)$
      \STATE $j\leftarrow\mathcal{U}\left(1,m-l\right)$
      \STATE subsequences $\leftarrow$ subsequences $\cup \left\{ \left(i,s,j,l\right) \right\}$
    \ENDFOR
  \ENDFOR
\ENDFOR
\STATE 
\STATE $\triangleright$ Generate the transformed dataset using $\mathcal{T}$ and the subsequences
\STATE $X\in\mathbb{R}^{n\times p}$
\FOR{$j=1$ to $p$}
  \STATE $\left(i',s,j',l\right)\leftarrow$ subsequences.get(j)
  \FOR{$i=1$ to $n$}
    \STATE $x_{i,j}\leftarrow dist\left(\left(t_{i',s,j'},\ldots t_{i',s,j'+l}\right),T_{i,s}\right)$
  \ENDFOR
\ENDFOR
\RETURN{$\left(X,Y\right)$}
\end{algorithmic}
\end{algorithm}

\subsection{Considering Derivatives of Time Series\label{sub:Derivatives-of-Time-Series}}

\begin{figure}
\begin{centering}
\includegraphics[height=5cm]{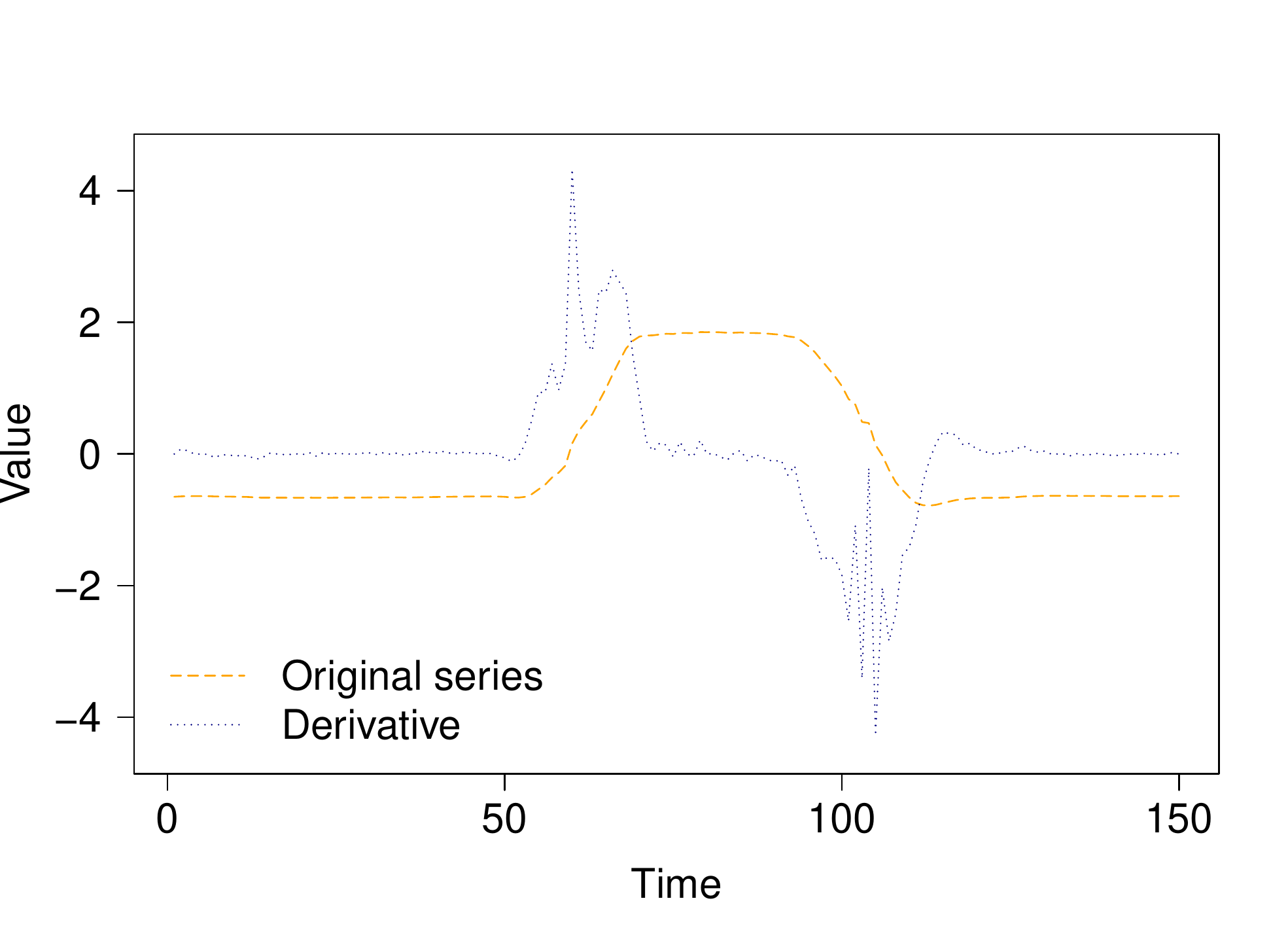}
\par\end{centering}

\caption{GunPoint time series instance and its time series derivative.\label{fig:Derivatives-of-time-series}}

\end{figure}
In the next section, Ultra-Fast Shapelets is compared to other state-of-the-art
methods for multivariate time series classification. To ensure a fair
comparison, the same classification model and the same features shall
be used. Since SMTS \cite{Baydogan2014} is not only using the raw
time series but also the derivative of it, it is now explained how
derivatives can be used for Ultra-Fast Shapelets.

The derivative of a time series $T=\left(t_{1},\ldots,t_{m}\right)$
is defined as $\nabla T=\left(0,t_{2}-t_{1},\ldots,t_{m}-t_{m-1}\right)$.
The leading $0$ ensures that the derivative has the same length as its time series.
The use of derivatives of time series is straight forward for Ultra-Fast Shapelets.
For each stream $T_{i}$ of a time series $T$ a new stream
$\nabla T_{i}$ is added which is the derivative of $T_{i}$. The
original dataset is now twice that large but no adaption to Algorithm
\ref{alg:Generalized-Shapelet-Sampling} needs to be done. An example
for a time series derivative is given in Figure \ref{fig:Derivatives-of-time-series}.

\section{Experiments\label{sec:Experiments}}

In Section \ref{sub:Baselines}, the baselines are summarized in order
to make this work self-contained. Section \ref{sub:Experimental-Setup}
provides a detailed description of the experimental setup, explaining
which hyperparameters are used and how they have been found for every
algorithm.

One claim is that sampling of shapelets is in general a scalable way
to extract features for time series classification which is also accurate.
Therefor, experiments are conducted on 52 univariate datasets from
various domains such as speech recognition, activity recognition,
medicine, image classification and several more. These datasets are
downloaded from \cite{Bagnall,Keogh} and have different properties
such as number of training instances, classes and length (see Table
\ref{tab:uts-dataset-stats}). In Section \ref{sub:Univariate-Datasets},
Ultra-Fast Shapelets (UFS) is compared to various shapelet-based classifiers
and its ability to compete is demonstrated. Furthermore, the runtime
is compared and it is shown that Ultra-Fast Shapelets is indead faster.

Finally, in Section \ref{sub:Multivariate-Datasets}, UFS is compared
to state-of-the-art classifiers for multivariate time series on 15
datasets from different domains such as speech, gesture, motion, handwriting
and sign language recogniton provided by \cite{Baydogan}. The number
of streams varies between 2 and 62, the number of classes between
2 and 95. In contrast to the univariate datasets, some of the multivariate
datasets do not have a fixed length per instance per dataset. A detailed
description of the datasets is given in Table \ref{tab:mts-dataset-stats}.

\subsection{Experimental Setup\label{sub:Experimental-Setup}}

Ultra-Fast Shapelets (UFS) has only a single hyperparameter that is the
percentage of considered candidates. Instead of tuning the percentage
hyperparameter carefully for all 67 datasets, the highest non-positive
power of 10 was chosen such that the number of expected candidates
is smaller than 10,000 per stream. For all univariate datasets the
normalized distance function defined in Equation \ref{eq:minNormDist}
was used, for the multivariate datasets this decision was added to
the hyperparameter search and then either the distance function defined
in Equation \ref{eq:minNormDist} or \ref{eq:minDist} was used.

On the transformed data any classifier may be trained. Random Forest
and a linear SVM were chosen because the results for SMTS are reported
using a Random Forest and Learning Shapelets is using a linear classifier.

A random forest has three hyperparameters, i.e. the number of trees
$J$, the number of sampled features $v$, and the depth $d$ of each
tree. A gridsearch was applied on $J\in\left\{ 20,50,100,500,1000\right\} $,
$v\in\left\{ \left\lfloor \log\left(p+1\right)\right\rfloor ,p\right\} $
and $d\in\left\{ 5,\ldots,2\log\left(p\right)\right\} $ and evaluated
against the out-of-bag error (OOE). The model with smallest OOE
was used to classify the test instances. In all experiments
the mean over ten repetitions is reported. In case that there was
more than one model with the smallest OOE, one was taken at random.

A linear SVM has usually only one hyperparameter, i.e. the regularization
term $C$. Another hyperparameter was added that decides whether to
use a $L1$ or $L2$-regularized SVM. A gridsearch was applied on
$C\in\left\{ 2^{1},\ldots,2^{10}\right\} $ and both regularization
methods. The best combination in a 5-fold cross-validation was used
to report the error on test. Again, in all experiments the mean
over ten replications is reported.

The results of SMTS, nearest neighbor with dynamic time warping distance
without a warping window (NNDTW) and a multivariate extension of TSBF
\cite{Baydogan2013} (MTSBF) were taken from Baydogan et al. \cite{Baydogan2014}.
Results for the other baselines were redone since the authors only
report the accuracy on one run and otherwise the runtime comparison
would have not been fair. For the experiments, the implementation
provided by the original authors were taken and the hyperparameters were used
as proposed by the authors.
Because of this and the fact that some datasets are too large, not every baseline is
evaluated on every dataset but this shall ensure that the best hyperparameters
are chosen.
Again, the average over ten replications is reported for
every dataset.

\subsection{Univariate Datasets\label{sub:Univariate-Datasets}}

To support the claim that Ultra-Fast Shapelets is comparably to state-of-the-art
shapelet-based classifiers with respect to accuracy but faster, experiments
are conducted on 52 univariate time series datasets and compared to
three different methods of shapelet discovery using various classifiers.
Ultra-Fast Shapelets (UFS) using Random Forest (RF) and a linear SVM (SVM)
is compared to the exhaustive search of shapelets (ES) using RF and
SVM, to Fast Shapelets (FS) and to Learning Shapelets (LS). These
four shapelet-based methods are compared based on classification
error in Section \ref{sub:Classification-Accuracy-UTS} and runtime
in Section \ref{sub:Runtime-Analysis-UTS}. The experiments show
that UFS is as good as state-of-the-art shapelet-based classifiers
but scale to large data and therefor can be used for both, univariate
and multivariate time series classification, with decent accuracy
and runtime. More detailed information about the baselines is given
in Section \ref{sub:Baselines}.

The datasets are provided by the UCR time series database \cite{Keogh}
and by Bagnall et al. \cite{Bagnall}. Table \ref{tab:uts-dataset-stats}
contains few statistics about the datasets, further information can
be found on the corresponding websites \cite{Bagnall,Keogh}.

\begin{table}
\centering
\caption{Characteristics of the univariate datasets.\label{tab:uts-dataset-stats}}
\resizebox{!}{.32\paperheight}{
\begin{tabular}{llllll}
\hline\noalign{\smallskip}
Dataset & Train Instances & Test Instances & Length & Classes & Candidates\\
\noalign{\smallskip}\hline\noalign{\smallskip}
50words & 450 & 455 & 270 & 50 & 16,220,700\\
Adiac & 390 & 391 & 176 & 37 & 5,937,750\\
Beef & 30 & 30 & 470 & 5 & 3,292,380\\
CBF & 30 & 900 & 128 & 3 & 240,030\\
ChlorineConcentration & 467 & 3840 & 166 & 3 & 6,318,510\\
CinC\_ECG\_torso & 40 & 1380 & 1639 & 4 & 53,628,120\\
Coffee & 28 & 28 & 286 & 2 & 1,133,160\\
Cricket\_X & 390 & 390 & 300 & 12 & 17,374,890\\
Cricket\_Y & 390 & 390 & 300 & 12 & 17,374,890\\
Cricket\_Z & 390 & 390 & 300 & 12 & 17,374,890\\
DiatomSizeReduction & 16 & 306 & 345 & 4 & 943,936\\
DP\_Little & 400 & 645 & 250 & 3 & 12,350,400\\
DP\_Middle & 400 & 645 & 250 & 3 & 12,350,400\\
DP\_Thumb & 400 & 645 & 250 & 3 & 12,350,400\\
ECG200 & 100 & 100 & 96 & 2 & 446,500\\
ECGFiveDays & 23 & 861 & 136 & 2 & 208,035\\
FaceAll & 560 & 1690 & 131 & 14 & 4,695,600\\
FaceFour & 24 & 88 & 350 & 4 & 1,457,424\\
FacesUCR & 200 & 2050 & 131 & 14 & 1,677,000\\
FISH & 175 & 175 & 463 & 7 & 18,635,925\\
Gun\_Point & 50 & 150 & 150 & 2 & 551,300\\
Haptics & 155 & 308 & 1092 & 5 & 92,162,225\\
InlineSkate & 100 & 550 & 1882 & 7 & 176,814,000\\
ItalyPowerDemand & 67 & 1029 & 24 & 2 & 16,951\\
Lighting2 & 60 & 61 & 637 & 2 & 12,115,800\\
Lighting7 & 70 & 73 & 319 & 7 & 3,528,210\\
MALLAT & 55 & 2345 & 1024 & 8 & 28,751,415\\
MedicalImages & 381 & 760 & 99 & 10 & 1,810,893\\
MoteStrain & 20 & 1252 & 84 & 2 & 68,060\\
MP\_Little & 400 & 645 & 250 & 3 & 12,350,400\\
MP\_Middle & 400 & 645 & 250 & 3 & 12,350,400\\
OliveOil & 30 & 30 & 570 & 4 & 4,847,880\\
OSULeaf & 200 & 242 & 427 & 6 & 18,105,000\\
Otoliths & 64 & 64 & 512 & 2 & 8,339,520\\
PP\_Little & 400 & 645 & 250 & 3 & 12,350,400\\
PP\_Middle & 400 & 645 & 250 & 3 & 12,350,400\\
PP\_Thumb & 400 & 645 & 250 & 3 & 12,350,400\\
SonyAIBORobotSurface & 20 & 601 & 70 & 2 & 46,920\\
SonyAIBORobotSurfaceII & 27 & 953 & 65 & 2 & 54,432\\
StarLightCurves & 1000 & 8236 & 1024 & 3 & 522,753,000\\
SwedishLeaf & 500 & 625 & 128 & 15 & 4,000,500\\
Symbols & 25 & 995 & 398 & 6 & 1,965,150\\
synthetic\_control & 300 & 300 & 60 & 6 & 513,300\\
Trace & 100 & 100 & 275 & 4 & 3,740,100\\
TwoLeadECG & 23 & 1139 & 82 & 2 & 74,520\\
Two\_Patterns & 1000 & 4000 & 128 & 4 & 8,001,000\\
uWaveGestureLibrary\_X & 896 & 3582 & 315 & 8 & 44,030,336\\
uWaveGestureLibrary\_Y & 896 & 3582 & 315 & 8 & 44,030,336\\
uWaveGestureLibrary\_Z & 896 & 3582 & 315 & 8 & 44,030,336\\
wafer & 1000 & 6164 & 152 & 2 & 11,325,000\\
WordsSynonyms & 267 & 638 & 270 & 25 & 9,624,282\\
yoga & 300 & 3000 & 426 & 2 & 27,030,000\\
\noalign{\smallskip}\hline
\end{tabular}
}
\end{table}

\subsubsection{Classification Accuracy\label{sub:Classification-Accuracy-UTS}}

In this section the results of an empirical comparison on 52 datasets
of various domains for different methods of shapelet discovery are
presented. Detailed results for each dataset are shown in Table \ref{tab:uts-acc}.

The different methods can be divided into two groups depending on
which kind of classifier they use. The methods that are using a linear
SVM (UFS (SVM), ES (SVM), LS) yield better results than those which are
using a non-linear, tree-based classifier. The tree-based classifiers
are only in 9 datasets better than the SVM which confirms the results
by Hills et al. \cite{Hills2014}. Overall, FS is the worst classifier which
is not surprising because it is an approximation of ES. Comparing
UFS with ES by classifier, the performance using a SVM is almost similar.
ES has better prediction quality on 10 datasets, UFS on 15. UFS using
a random forest shows better performance in 19 datasets and is worse
than ES in 7. These results show a strong empirical evidence that
simply choosing subsequences at random will not deteriorate the accuracy
in comparison to more informative methods like ES or FS.

While LS achieves better accuracy than UFS, it has a higher runtime performance
by an average of three orders of magnitude and is the slowest among all investigated methods, as shown in the next section.
In that context, the proposed method (UFS) is more scalable than LS for large datasets.
In terms of hyperparameters, LS requires three sensitive hyperparameters
while our proposed method has only a single insensitive hyperparameter (see Section \ref{sub:Hyperparameter-Sensitivity}).
This excludes the classifier-specific hyperparameters for both methods.
Finally, one of the reasons that LS is more accurate is that it is not limited
in the number of candidates. While ES, FS and UFS can only choose among
subsequences that appear in the training set, LS can choose any subsequence.

\begin{sidewaystable}
\caption{Test error rates averaged over 10 replications for different univariate time series classification methods compared to Ultra-Fast Shapelets on 52 datasets. \label{tab:uts-acc}}
\begin{tabular}{lp{1cm}p{1cm}lp{1cm}p{1cm}l|lp{1cm}p{1cm}l}
\hline\noalign{\smallskip}
Dataset & UFS (SVM) & ES (SVM) & LS & UFS (RF) & ES (RF) & FS & Dataset & UFS (SVM) & UFS (RF) & FS\\
\noalign{\smallskip}\hline\noalign{\smallskip}
Adiac & \textbf{0.276} & 0.757 & 0.581 & 0.281 & 0.704 & 0.442 & 50words & \textbf{0.184} & 0.278 & 0.497\\
Beef & 0.263 & \textbf{0.100} & 0.193 & 0.423 & 0.360 & 0.483 & CBF & \textbf{0.002} & 0.014 & 0.062\\
ChlorineConcentration & \textbf{0.262} & 0.429 & 0.270 & 0.357 & 0.338 & 0.422 & CinC\_ECG\_torso & \textbf{0.151} & 0.199 & 0.274\\
Coffee & 0.011 & \textbf{0.000} & \textbf{0.000} & 0.054 & 0.061 & 0.075 & Cricket\_X & \textbf{0.225} & 0.294 & 0.538\\
DiatomSizeReduction & 0.075 & 0.082 & \textbf{0.058} & 0.084 & 0.218 & 0.131 & Cricket\_Y & \textbf{0.234} & 0.279 & 0.506\\
DP\_Little & 0.293 & 0.323 & \textbf{0.257} & 0.315 & 0.364 & 0.434 & Cricket\_Z & \textbf{0.207} & 0.248 & 0.561\\
DP\_Middle & 0.288 & 0.305 & \textbf{0.271} & 0.288 & 0.384 & 0.419 & ECG200 & \textbf{0.119} & 0.187 & 0.230\\
DP\_Thumb & 0.291 & 0.304 & \textbf{0.264} & 0.302 & 0.366 & 0.420 & FaceAll & \textbf{0.212} & 0.260 & 0.366\\
ECGFiveDays & \textbf{0.000} & 0.001 & \textbf{0.000} & 0.007 & 0.013 & 0.004 & FacesUCR & \textbf{0.043} & 0.090 & 0.310\\
FaceFour & 0.039 & 0.023 & \textbf{0.019} & 0.043 & 0.226 & 0.092 & FISH & \textbf{0.035} & 0.105 & 0.180\\
Gun\_Point & 0.021 & \textbf{0.000} & 0.006 & 0.011 & 0.046 & 0.068 & Haptics & \textbf{0.509} & 0.529 & 0.611\\
ItalyPowerDemand & 0.038 & 0.065 & \textbf{0.036} & 0.052 & 0.072 & 0.079 & InlineSkates & 0.618 & \textbf{0.573} & 0.738\\
Lighting7 & \textbf{0.222} & 0.342 & 0.370 & 0.264 & 0.363 & 0.399 & Lighting2 & \textbf{0.213} & 0.236 & 0.316\\
MedicalImages & \textbf{0.260} & 0.477 & 0.336 & 0.300 & 0.517 & 0.392 & MALLAT & 0.044 & \textbf{0.021} & 0.058\\
MoteStrain & 0.134 & 0.119 & \textbf{0.099} & 0.131 & 0.148 & 0.216 & OliveOil & \textbf{0.110} & 0.123 & 0.292\\
MP\_Little & 0.283 & 0.293 & \textbf{0.263} & 0.268 & 0.336 & 0.435 & OSULeaf & \textbf{0.158} & 0.245 & 0.323\\
MP\_Middle & 0.252 & \textbf{0.234} & 0.241 & 0.235 & 0.303 & 0.395 & SonyAIBORobotSurfaceII & \textbf{0.088} & 0.121 & 0.209\\
Otoliths & 0.434 & 0.359 & \textbf{0.319} & 0.436 & 0.364 & 0.440 & StarLightCurves & 0.031 & \textbf{0.029} & 0.054\\
PP\_Little & 0.316 & \textbf{0.279} & 0.297 & 0.340 & 0.328 & 0.451 & SwedishLeaf & \textbf{0.065} & 0.094 & 0.225\\
PP\_Middle & \textbf{0.256} & 0.287 & 0.258 & 0.263 & 0.329 & 0.420 & Two\_Patterns & 0.003 & \textbf{0.001} & 0.079\\
PP\_Thumb & \textbf{0.304} & 0.309 & \textbf{0.304} & 0.342 & 0.391 & 0.465 & uWaveGestureLibrary\_X & 0.197 & \textbf{0.191} & 0.297\\
SonyAIBORobotSurface & 0.122 & \textbf{0.060} & 0.090 & 0.109 & 0.080 & 0.302 & uWaveGestureLibrary\_Y & 0.291 & \textbf{0.284} & 0.386\\
Symbols & 0.111 & 0.167 & 0.069 & \textbf{0.067} & 0.176 & 0.070 & uWaveGestureLibrary\_Z & 0.271 & \textbf{0.238} & 0.363\\
synthetic\_control & \textbf{0.006} & 0.097 & 0.021 & 0.009 & 0.081 & 0.083 & wafer & 0.003 & 0.008 & \textbf{0.002}\\
Trace & \textbf{0.000} & \textbf{0.000} & \textbf{0.000} & 0.008 & \textbf{0.000} & \textbf{0.000} & WordsSynonyms & \textbf{0.269} & 0.376 & 0.569\\
TwoLeadECG & 0.025 & \textbf{0.000} & 0.002 & 0.075 & 0.064 & 0.741 & yoga & \textbf{0.129} & 0.150 & 0.294\\
\noalign{\smallskip}\hline
\end{tabular}
\end{sidewaystable}


\subsubsection{Runtime Analysis\label{sub:Runtime-Analysis-UTS}}

This section shows that Ultra-Fast Shapelets (UFS) is not only as accurate
as state-of-the-art methods but also significantly faster which makes
it finally feasible to apply shapelets on multivariate time series
datasets. Therefor, the four shapelet-based methods are compared empirically
and theoretically.

Starting with the empirical runtime analysis, Table \ref{tab:uts-time}
provides the measured runtime in seconds
averaged over 10 replications for each method on 52 datasets. The subsequence lengths
that are considered are those proposed by the authors. This means
that only UFS and Fast Shapelets (FS) are considering all candidates
while the exhaustive search (ES) and Learning Shapelets (LS) only
consider a subset of them. Knowing this, note that this is obviously
an advantage for ES and LS. Nevertheless, FS and UFS are significantly
faster. UFS is slow compared to the baselines for the small datasets
but in general faster which means that it is faster than the fastest
shapelet discovery method so far (FS) in 45 out of 52 datasets, in some
even by a factor of 100.

The number of all subsequences respectively candidates in a time series
dataset with $n$ instances of length $m$ is $c=\mathcal{O}\left(nm^{2}\right)$.
Since the exhaustive search compares each possible pair of candidates
of equal length, the runtime is $\mathcal{O}\left(n^{2}m^{4}\right)$.
FS reduces the number of candidates to a subset $r<c$ which is then
compared with all $\mathcal{O}\left(nm^{2}\right)$ candidates of
equal length so that FS needs time $\mathcal{O}\left(rnm^{2}\right)$.
LS \cite{Grabocka2014} reports a runtime of $\mathcal{O}\left(ipnm^{2}\right)$
where $i$ is the number of iterations used until the algorithm converges
and $p$ is the number of shapelets that have to be found. Since the
authors propose a very high number of iterations and shapelets for
some datasets, a comparably high runtime is observed. Finally, UFS
chooses randomly a subset of $p$ candidates, where for the here executed
experiments $p<10,000$. Thus, the runtime is $\mathcal{O}\left(pnm^{2}\right)$
and $p<r$ for the larger datasets but $p\approx nm^{2}$ for the
very small datasets which explains the high runtime for those. Since
the number of candidates $p$ is not optimized but a rule of thumb
was used, one can further improve the speed without a loss of accuracy
as shown in Section \ref{sub:Hyperparameter-Sensitivity}.
Theoretical and empirical results are summarized in Table \ref{fig:Summary-of-runtime-results}.
\begin{table}
\caption{Summary of theoretical and empirical runtime results. (win/lose)\label{fig:Summary-of-runtime-results}}
\begin{tabular}{llllll}
\hline\noalign{\smallskip}
 & UFS & FS & LS & ES\\
\noalign{\smallskip}\hline\noalign{\smallskip}
Runtime complexity & $\mathcal{O}\left(pnm^{2}\right)$ & $\mathcal{O}\left(rnm^{2}\right)$ & $\mathcal{O}\left(ipnm^{2}\right)$ & $\mathcal{O}\left(n^{2}m^{4}\right)$\\
UFS & - & 45/7 & 26/0 & 23/3\\
FS & 7/45 & - & 26/0 & 26/0\\ 
LS & 0/26 & 0/26 & - & 11/15\\
ES & 3/23 & 0/26 & 15/11 & -\\
\noalign{\smallskip}\hline
\end{tabular}
\end{table}

\begin{sidewaystable}
\caption{Measured time in seconds for learning the model and discovering the shapelets averaged over 10 replications for different shapelet-based univariate time series classification methods compared to Ultra-Fast Shapelets on 52 datasets. \label{tab:uts-time}}
\begin{tabular}{llllll|lll}
\hline\noalign{\smallskip}
Dataset & UFS & FS & LS & ES & Dataset & UFS & FS\\
\noalign{\smallskip}\hline\noalign{\smallskip}
Adiac & \textbf{45.8} & 332.6 & 203847.0 & 5522.6 & 50words & \textbf{23.2} & 2198.1\\
Beef & \textbf{8.3} & 196.9 & 25063.5 & 2148.0 & CBF & \textbf{8.7} & 10.9\\
ChlorineConcentration & \textbf{271.9} & 760.3 & 30743.7 & 13756.5 & CinC\_ECG\_torso & \textbf{3201.6} & 4398.9\\
Coffee & \textbf{1.6} & 22.5 & 372.6 & 204.7 & Cricket\_X & \textbf{25.7} & 3756.0\\
DiatomSizeReduction & 63.4 & \textbf{15.6} & 14184.1 & 94.4 & Cricket\_Y & \textbf{26.6} & 3605.7\\
DP\_Little & \textbf{17.1} & 975.1 & 8384.8 & 66570.1 & Cricket\_Z & \textbf{25.9} & 4679.2\\
DP\_Middle & \textbf{17.3} & 996.1 & 23875.4 & 143041.5 & ECG200 & \textbf{2.7} & 16.3\\
DP\_Thumb & \textbf{17.9} & 916.3 & 9107.0 & 109388.6 & FaceAll & \textbf{58.8} & 757.5\\
ECGFiveDays & 10.9 & \textbf{3.6} & 59.2 & 122.0 & FacesUCR & \textbf{19.1} & 280.3\\
FaceFour & \textbf{5.6} & 102.9 & 11078.3 & 4384.8 & FISH & \textbf{244.4} & 935.6\\
Gun\_Point & \textbf{0.7} & 9.5 & 1475.3 & 474.1 & Haptics & \textbf{818.0} & 12491.0\\
ItalyPowerDemand & 1.8 & \textbf{0.4} & 20.8 & 1.5 & InlineSkate & \textbf{637.0} & 22677.2\\
Lighting7 & \textbf{10.1} & 322.8 & 3902.6 & 13409.5 & Lighting2 & \textbf{11.1} & 1131.3\\
MedicalImages & \textbf{7.4} & 371.5 & 6457.7 & 11084.6 & MALLAT & \textbf{121.7} & 1736.5\\
MoteStrain & 17.2 & \textbf{3.1} & 410.5 & 7.8 & OliveOil & \textbf{16.8} & 107.2\\
MP\_Little & \textbf{18.1} & 996.5 & 55964.1 & 75666.3 & OSULeaf & \textbf{26.5} & 1629.7\\
MP\_Middle & \textbf{17.0} & 993.3 & 6797.5 & 117056.5 & SonyAIBORobotSurfaceII & 8.4 & \textbf{1.3}\\
Otoliths & \textbf{47.5} & 303.2 & 9453.2 & 46546.0 & StarLightCurves & \textbf{8278.9} & 21473.5\\
PP\_Little & \textbf{17.3} & 918.8 & 27699.3 & 67944.0 & SwedishLeaf & \textbf{23.5} & 451.7\\
PP\_Middle & \textbf{17.0} & 950.2 & 35425.6 & 52285.4 & Two\_Patterns & \textbf{147.0} & 957.2\\
PP\_Thumb & \textbf{17.5} & 943.7 & 67798.2 & 77402.3 & uWaveGestureLibrary\_X & \textbf{344.6} & 4827.5\\
SonyAIBORobotSurface & 4.9 & \textbf{1.1} & 115.0 & 8.5 & uWaveGestureLibrary\_Y & \textbf{348.9} & 4379.6\\
Symbols & \textbf{52.0} & 93.0 & 1104.3 & 12511.4 & uWaveGestureLibrary\_Z & \textbf{497.0} & 5215.9\\
synthetic\_control & \textbf{5.6} & 63.9 & 783.3 & 1285.2 & wafer & \textbf{56.6} & 190.5\\
Trace & \textbf{11.2} & 181.0 & 11561.9 & 47665.3 & WordsSynonyms & \textbf{121.8} & 1140.0\\
TwoLeadECG & 16.3 & \textbf{1.3} & 45.6 & 3.5 & yoga & \textbf{271.7} & 1711.6\\
\noalign{\smallskip}\hline
\end{tabular}
\end{sidewaystable}

\subsection{Hyperparameter Sensitivity\label{sub:Hyperparameter-Sensitivity}}

\begin{figure}
\begin{centering}
\includegraphics[width=1\textwidth]{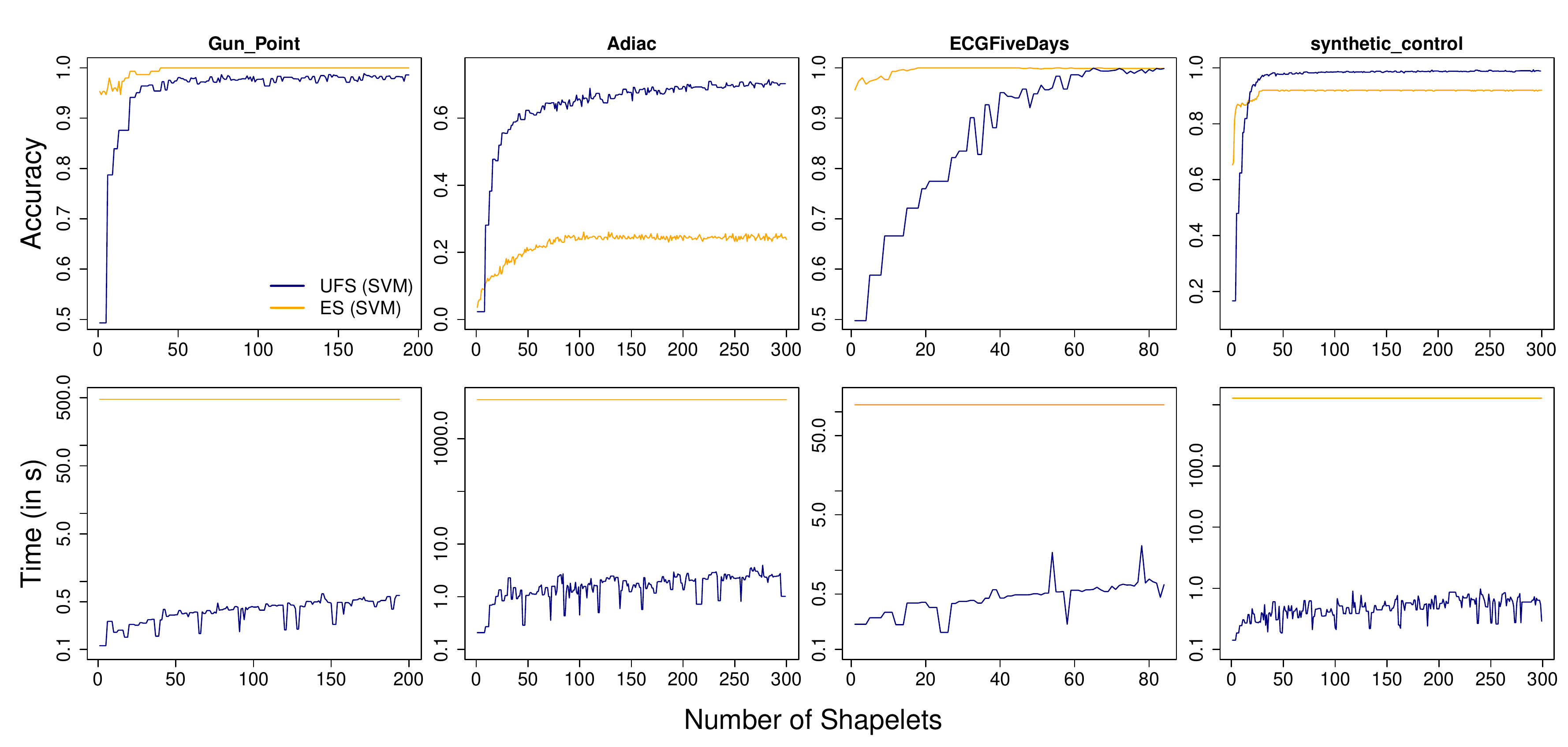}
\par\end{centering}

\caption{Accuracy achieved and time needed on four different datasets for the
exhaustive search (orange) and Ultra-Fast Shapelets (blue), both using
a linear SVM for classification.\label{fig:Varying-number-of-shapelets}}

\end{figure}

Ultra-Fast Shapelets has only a single hyperparameter, i.e. the number
of chosen subsequences plus those needed for the chosen classifier.
This section is devoted to two questions: i) is Ultra-Fast Shapelets
sensitive to the hyperparameter and is it easy to find an optimal
one and ii) is Ultra-Fast Shapelets better because it uses more features
respectively does it help the baselines if they also use more features.
These two question will be answered exemplarily at four random datasets.
Figure \ref{fig:Varying-number-of-shapelets} shows the accuracy and
the time needed for Ultra-Fast Shapelets (UFS) and the exhaustive search
(ES) using a linear SVM. The measured time contains the time for discovering
the shapelets as well as training the model.

First of all, it seems that, even if ES uses as many features as UFS,
the accuracy does not improve. Actually, with increasing number of
shapelets, the accuracy improves until it converges at some point
which makes it easy to find the best number of shapelets for both
methods. ES profits from the supervised features selection and tends
to need less features to achieve its best accuracy. The random subsequence
selection of UFS tends to result in the need of more features but needs
orders of magnitudes less time to find them. Finally, if the results
for UFS in Figure \ref{fig:Varying-number-of-shapelets} are compared
to the reported results in Table \ref{tab:uts-acc} and
\ref{tab:uts-time} it is clear that a hyperparameter search
will further improve the runtime without harming the accuracy.

\subsection{Multivariate Datasets\label{sub:Multivariate-Datasets}}

The 15 multivariate datasets used to evaluate Ultra-Fast Shapelets are
from various domains such as sign language recognition (AUSLAN), handwriting
recognition (CharacterTrajectories, PenDigits), motion (CMU\_MOCAP\_S16),
gesture (uWaveGestureLibrary) and speech recognition (ArabicDigits,
JapaneseVowels). Detailed characteristics of the datasets are given
in Table \ref{tab:mts-dataset-stats}, detailed results are presented
in Table \ref{tab:mts-acc}. A summary of the empirical evaluation
is given in Table \ref{fig:Summary-of-empirical-evaluation-mts}.

There exist two variations of Ultra-Fast Shapelets: Ultra-Fast Shapelets
with ($\nabla$UFS) and without time series derivatives (UFS). Time
series derivatives were added since Symbolic Representation for Multivariate
Timeseries (SMTS) are using them by default. For SMTS it was shown
that this additional information improves the accuracy. Time series
derivatives can improve the accuracy for UFS in some cases. In some
they do not help, in some they may deteriorate the accuracy since
they are adding less helpful predictors. Nevertheless, UFS with derivatives
is in 11 out of 15 cases better than without derivatives.

UFS with and without derivatives outperforms the baselines. $\nabla$UFS
is for 11, UFS for 10 out of 15 datasets better than SMTS. For NNDTW
the number of cases is even 14 or 13, respectively. Comparing UFS
to MTSBF, UFS is better in 5 or 6 out of 7 datasets, depending whether
time series derivatives are used or not.

Concluding, UFS with derivatives is more accurate than without. UFS
is better than SMTS even if no time series derivatives are used and
MTSBF has a similar prediction accuracy as SMTS. NNDTW is the worst
of the compared classification methods.

\begin{table}
\caption{Characteristics of the multivariate datasets.\label{tab:mts-dataset-stats}}
\begin{tabular}{llllll}
\hline\noalign{\smallskip} 
Dataset & Train Instances & Test Instances & Streams & Length & Classes\\
\noalign{\smallskip}\hline\noalign{\smallskip}
ArabicDigits & 6600 & 2200 & 13 & 4-93 & 10\\
AUSLAN & 1140 & 1425 & 22 & 45-136 & 95\\
CharacterTrajectories & 300 & 2558 & 3 & 109-205 & 20\\
CMU\_MOCAP\_S16 & 29 & 29 & 62 & 127-580 & 2\\
ECG & 100 & 100 & 2 & 39-152 & 2\\
JapaneseVowels & 270 & 370 & 12 & 7-29 & 9\\
Libras & 180 & 180 & 2 & 45 & 15\\
LP1 & 38 & 50 & 6 & 15 & 4\\
LP2 & 17 & 30 & 6 & 15 & 5\\
LP3 & 17 & 30 & 6 & 15 & 4\\
LP4 & 42 & 75 & 6 & 15 & 3\\
LP5 & 64 & 100 & 6 & 15 & 5\\
PenDigits & 300 & 10692 & 2 & 8 & 10\\
uWaveGestureLibrary & 200 & 4278 & 3 & 315 & 8\\
Wafer & 298 & 896 & 6 & 104-198 & 2\\
\noalign{\smallskip}\hline
\end{tabular}
\end{table}  

\begin{table}
\caption{Summary of empirical evaluation on multivariate time series. (win/lose)\label{fig:Summary-of-empirical-evaluation-mts}}
\begin{tabular}{llllll}
\hline\noalign{\smallskip} 
 & $\nabla$UFS (RF) & UFS (RF) & SMTS & NNDTW & MTSBF\\
\noalign{\smallskip}\hline\noalign{\smallskip}
$\nabla$UFS (RF) & - & \textbf{11/2} & \textbf{11/4} & \textbf{14/1} & \textbf{6/1}\\
UFS (RF) & 2/11 & - & \textbf{10/4} & \textbf{13/1} & \textbf{5/2}\\
SMTS & 4/11 & 4/10 & - & \textbf{13/2} & 3/3\\
NNDTW & 1/14 & 1/13 & 2/13 & - & 2/5\\
MTSBF & 1/6 & 2/5 & 3/3 & \textbf{5/2} & -\\
\noalign{\smallskip}\hline
\end{tabular}
\end{table}

\begin{table}
\caption{Test error rates averaged over 10 replications for different multivariate time series classification methods compared to Ultra-Fast Shapelets. \label{tab:mts-acc}}
\begin{tabular}{llllll} 
\hline\noalign{\smallskip} 
Dataset & $\nabla$UFS (RF) & UFS (RF) & SMTS& NNDTW & MTSBF\\
\noalign{\smallskip}\hline\noalign{\smallskip}
ArabicDigits & \textbf{0.033} & 0.036 & 0.036 & 0.092 & -\\
AUSLAN & 0.021 & 0.028 & 0.053 & 0.238 & \textbf{0.000}\\
CharacterTrajectories & \textbf{0.007} & \textbf{0.007} & 0.008 & 0.040 & 0.033\\
CMU\_MOCAP\_S16 & \textbf{0.000} & \textbf{0.000} & 0.003 & 0.069 & 0.003\\
ECG & 0.151 & \textbf{0.138} & 0.182 & 0.150 & 0.165\\
JapaneseVowels & 0.056 & 0.068 & \textbf{0.031} & 0.351 & -\\
Libras & 0.111 & 0.151 & \textbf{0.091} & 0.200 & 0.183\\
LP1 & \textbf{0.042} & 0.058 & 0.144 & 0.280 & -\\
LP2 & 0.293 & 0.307 & \textbf{0.240} & 0.467 & -\\
LP3 & 0.213 & \textbf{0.207} & 0.240 & 0.500 & -\\
LP4 & \textbf{0.068} & 0.085 & 0.105 & 0.187 & -\\
LP5 & \textbf{0.287} & 0.296 & 0.349 & 0.480 & -\\
PenDigits & \textbf{0.073} & 0.081 & 0.083 & 0.088 & -\\
uWaveGestureLibrary & 0.061 & 0.071 & \textbf{0.059} & 0.071 & 0.101\\
Wafer & \textbf{0.013} & 0.024 & 0.035 & 0.023 & 0.015\\
\noalign{\smallskip}\hline\noalign{\smallskip}
Wins & 8 & 4 & 4 & 0 & 1\\
\noalign{\smallskip}\hline
\end{tabular} 
\end{table}

\section{Conclusions\label{sec:Conclusions}}

We proposed an ultra-fast way of extracting shapelets motivated by the
knowledge of redundant subsequences. Because the shapelet discovery
by most authors so far is nothing but a feature subset selection 
which is costly in time, it was not surprising that our method,
Ultra-Fast Shapelets, reduces the runtime by order of magnitudes
and is, to the best of our knowledge, the fastest so far published
shapelet discovery method.

Furthermore, the ultra-fast shapelet discovery method enabled us
to apply shapelet-based classifiers on long univariate datasets as well
as on multivariate time series. We compared UFS on 52 univariate datasets
with current state-of-the-art shapelet-based methods and showed empirically that it is competitive
in terms of accuracy.
Additionally, a comparison to state-of-the-art methods for multivariate
time series classification on 15 datasets from various domains has
shown that Ultra-Fast Shapelets creates predictive features. A Random
Forest classifier was better with shapelet features in 11 out of 15
cases compared to SMTS features.



\section{Acknowledgements}
Partially co-funded by the Seventh Framework Programme of the European Comission, through project REDUCTION (\# 288254). \url{www.reduction-project.eu}


\bibliography{bibtex}

\end{document}